\documentclass[a4paper,fleqn]{cas-sc}

\usepackage[numbers]{natbib}
\usepackage[utf8]{inputenc}
\DeclareUnicodeCharacter{2061}{}
\usepackage{amsmath}

\usepackage{fancyhdr}
\usepackage{multirow}
\usepackage{rotating}
\usepackage{placeins}
\usepackage{caption}
\usepackage{subcaption}
\def\tsc#1{\csdef{#1}{\textsc{\lowercase{#1}}\xspace}}
\tsc{WGM}
\tsc{QE}
\tsc{EP}
\tsc{PMS}
\tsc{BEC}
\tsc{DE}

\begin{document}
\let\WriteBookmarks\relax
\def\floatpagepagefraction{1}
\def\textpagefraction{.001}
\shorttitle{Plant Disease Detection using R-CNN}
\shortauthors{H. Rehana et~al.}

\title [mode = title]{Plant Disease Detection using Region-Based Convolutional Neural Network}

\author[1]{Hasin Rehana} [orcid=0000-0003-2992-6547]
\address[1]{Dept. of Computer Science and Engineering, University of Dhaka, Bangladesh}
\ead{hasin.cse13@gmail.com}
\author[1]{Muhammad Ibrahim*} [orcid=0000-0003-3284-8535]
\cortext[1]{Corresponding author}
\ead{ibrahim313@du.ac.bd}
\author[1]{Md. Haider Ali} [orcid=0000-0002-2629-2510]
\ead{haider@du.ac.bd}


\begin{abstract}
Agriculture plays an important role in the food and economy of Bangladesh. The rapid growth of population over the years also has increased the demand for food production. One of the major reasons behind low crop production is numerous bacteria, virus and fungal plant diseases. Early detection of plant diseases and proper usage of pesticides and fertilizers are vital for preventing the diseases and boost the yield. Most of the farmers use generalized pesticides and fertilizers in the entire fields without specifically knowing the condition of the plants. Thus the production cost oftentimes increases, and, not only that, sometimes this becomes detrimental to the yield. Deep Learning models are found to be very effective to automatically detect plant diseases from images of plants, thereby reducing the need for human specialists. This paper aims at building a lightweight deep learning model for predicting leaf disease in tomato plants.  By modifying the region-based convolutional neural network, we design an efficient and effective model that demonstrates satisfactory performance on a benchmark dataset. Our proposed model can easily be deployed in a larger system where drones take images of leaves and these images will be fed into our model to know the health condition.
\end{abstract}



\maketitle
\section{Introduction}

Bangladesh is an agriculture-based country with more than 75\% of the population depending on agriculture directly or indirectly \cite{a1}. Agricultural plants are usually prone to various diseases. Once caught, these diseases often spread out rapidly through an entire farm which results in a drastic economic loss for the farmers. So controlling the spread of the plant diseases at an earlier stage is very important to prevent such losses and to boost the yield.

When a plant is affected by a disease, oftentimes it is delineated in visually notable change in the texture and color of their leaves. 
Usually, each disease causes distinct changes in shape, color and texture of leaf, stem or root \cite{a2}. Thus, the quality, color, and texture of leaves may be visually examined to determine the health of the plants \cite{a3, a4}. 




Conventionally, identification of plant diseases is mainly performed either by lab experiments, or by the visual inspection conducted by agriculturists. However, these methods are quite infeasible in a developing country like Bangladesh due to the acute shortage of expert manpower in the rural area. 
In general, young farmers with less experience in agriculture are prone to making wrong decision in detecting diverse plant diseases and thus fail to identify whether the pesticides are needed or not. As a result, most of the farmers use generalized pesticides and fertilizers in their entire fields without specifically knowing the actual condition of the plants. However, not all insects are harmful, but the pesticides kill, along with harmful insects, these beneficial insects as well. 
Therefore, a smarter approach to identification of plant diseases is warranted.

\subsection{Motivation}

Recent advancement in computer vision and artificial intelligence, specifically in machine learning has facilitated automatic crop disease identification from images captured by bare smartphone cameras \cite{strawberry}. Over the last few years a few researches are performed to detect diseases of crops cultivated in Bangladesh like paddy, jute and wheat. 
However, after studying the relevant existing literature, we think that there is still room for improvement in these works, especially when it comes to the utilization of deep learning methods. That is why, this research focuses on the localization, detection as well as classification of plant leaf diseases using region-based CNN. Thus, the need for specialists to detect plant diseases in the crop field can be assisted by these automated systems.

As compared to many other domains such as healthcare, education, and entertainment, not too many research works are carried out that utilize computational techniques for agriculture. 
Among the existing machine learning systems used to detect plant leaf diseases, most of them are not suitable to be mass-used  by the root level farmers due to not being  efficient enough in terms of required computational resources. That is, these systems involve learning huge number of parameters which makes them inappropriate to be used in day-to-day smartphones used by rural farmers. Therefore, there is scope to develop a plant leaf disease detection model which will be lightweight and at the same time will detect the plant diseases effectively. \\

In particular, the objectives of this investigation are as follows:
\begin{itemize}
\item To develop an efficient, effective, and lightweight deep learning model for plant disease detection from leaf images.
\item To thoroughly analyze various aspects related to effectiveness and efficiency of our proposed model.
\item To compare the relative performance of the proposed model with the existing ones.

\end{itemize}

\subsection{Contributions}
We propose a modified region-based CNN. Specifically, we improve the detector network of Faster Region-Based CNN algorithm, known as Faster-RCNN \cite{a21}. 
We introduce a lightweight detector network by reducing the Fully Connected (FC) layers and the number of neurons in the remaining fully connected layer. We replace a fully connected layer having 4096 neurons by two inception modules in order to reduce the density. We also introduce parallel convolution with multiple shaped filters. Besides, the number of fully connected neurons in the remaining FC layer is also reduced to one-fourth to decrease the computation complexity and the chances of overfitting. We also modify the layers of RPN \cite{a21}. We apply bounding box optimization by finding the best Interest over Union (IoU) threshold. Our proposed model achieves 96\% accuracy on a benchmark dataset having around 10000 instances and outperforms two baseline models, namely Fast RCNN and Faster R-CNN across a number of configurations.

The remainder of the paper is organized as follows. Section \ref{sec:lit_review} reports the literature review and background study. Section \ref{sec:proposed} describes the proposed framework. Section \ref{sec:result} discusses the dataset properties, experimental settings, and analyzes the results. Finally, Section \ref{sec:conclusion} summarizes the findings of the research with hints at further research scope.

\section{Literature Review}
\label{sec:lit_review}
This section briefly describes the existing research works carried out for various plant disease detection techniques using machine learning. It also discusses some background knowledge required for understanding rest of the paper. 
\subsection{Related Work}

In the past several years, with the availability of cheap electronic devices and digital cameras, automated plant disease identification systems using different machine learning algorithms has become commonplace. Since the amount of relevant literature on the topic in question is large, below we mainly discuss the relevant literature, namely plant disease detection from leaf images using deep learning models. 

Anand et~al. \cite{a6} apply k-means clustering for brinjal leaf disease recognition using HOG features. Zhang et~al. \cite{a5} propose a fusion of K-means clustering with PHOG in which results in 85.25\% average accuracy for a four-class cucumber leaf disease segmentation and classification. But these Machine Learning algorithms demand handcrafted feature extraction procedures that leads to low detection accuracy. Misclassification often occurs due to inappropriate feature extraction. That is why over the last several years, deep learning techniques, especially CNN (\cite{cnn}) have produced a massive breakthrough in the area of image processing and computer vision. CNN is very effective for object classification as it alleviates the humans from devising complex features by hand. With the continuous involvement of researchers worldwide, many architectures of the CNN model, for example AlexNet \cite{a8} and VGGNet \cite{a9}, DenseNet\cite{a10} etc., have been proposed and are successfully being used to solve complex image processing and classification problems. Below we discuss the literature that uses deep learning models for plant disease detection.

S. Sladojevic \cite{a12} presents Caffenet, a single GPU variation of Alexnet, for classifying 13 different classes of leaf diseases using a dataset of around 34000 images. 
The model achieves around 96.3\% accuracy after 100 iterations. However, the training time is extremely high. Besides, for a small number of training images the accuracy falls below satisfactory level.

Ferentinos \cite{ferentinos2018deep} develops specialized deep learning models on specific CNNs for plant disease detection from its leaf with 25 plant species and 58 distinct classes of plant leaf images containing both healthy and disease infected data.

Barbedo et~al. \cite{barbedo2019plant} apply transfer learning for multiple plant leaves such as coconut tree, Corn, coffee, soybean and sugarcane. The  uniqueness of the research is, rather than considering the whole leaf image they explore the use of spots and lesions. The average classification accuracy is around 82\%. However, their proposed method needs manual symptom segmentation which prevents full automation.

Arsenovic et~al \cite{arsenovic2019solving} present a two-stage neural network architecture that  achieves the highest accuracy of 93.67\% where various conditions are considered such as complex background and multiple disease detection in a single image. A new plant leaf image dataset is introduced considering parameters like various weather conditions and different angles for analysis, which aims to solve the limitations of prior research.

P. Jiang et~al \cite{a15} presents a novel model for plant disease detection that is based on SSD (Single Shot Multibox). The authors perform several data augmentation methods that are implemented to reduce the chance of overfitting. The paper introduces a modified pre-trained network VGG-INCEP (VGGNet with Inception module of GoogleNet) that may extract multiscale disease spots. Rainbow concatenation method is then integrated with R-SSD. This deep learning-based approach can identify five common types of apple leaf diseases automatically with 78.8\% precision. A specialty of the model is that it can detect multiple diseases in the same image as well as different sizes of the same images.

P. K. Shetty et~al \cite{a11} introduce a system based on deep CNN that detects one of four types of rice leaf diseases from a dataset having 5932 images. Different types of data augmentation techniques are used for increasing the size of the dataset. 
Various pre-trained models such as Alexnet, VGG16, and VGG19 are applied for feature extraction and then SVM is applied for classification. The classification accuracy is 98.38\%. 

Ramesh et~al \cite{a14} propose a rice leaf diseases detection technique based on deep neural network. For image segmentation, K means clustering is used. Color feature, standard deviation, homogeneity, energy, mean, correlation, and contrast are extracted from these segmented images. These features are then used as input for the neural network and the accuracy varies between 89\% - 93\% for different diseases. On the other hand, K-NN yields 82\%-88\% accuracy for different disease classes. 

X. Xie et~al \cite{a16} introduce a deep learning model, namely Faster RCNN for grape leaf disease detection using a dataset having 62286 images. They also use a double Region Proposal Network(RPN) to support multiscale feature extraction, and Inception-ResNet module as the backbone network. The mean Average Precision (mAP) for the model is 74.3\%. The authors report that using data augmentation increases the mAP to some extent.

J. Sun et~al \cite{a17} utilize RPN to detect corn leaf diseases. They use Single Shot Detector (SSD) as the detector network. A 5 layered fully convolutional RPN model is designed for the task of region proposal. Interest over Union (IoU) metric is used for bounding box regression. Experimental result shows that SSD method achieves 71\% mAP.

Sardogan et~al. \cite{a13} presents a fusion of CNN model with Learning Vector Quantization (LVQ) technique. LVQ is a powerful heuristic algorithm that combines competitive learning with supervised learning. In their architecture, the fully connected layer of the CNN is fed into the input layer of LVQ algorithm. The system is tested on a tomato leaf dataset with 500 images.  
After 300 epochs with a 0.1 learning rate classification accuracy for the model is 86\%. 

S. Zhao \cite{zhao2021tomato} introduces a multi-scale CNN network architecture where the structure mainly includes attenuation extraction modules and residual blocks. Here SE module is deeply integrated with ResNet-50 which establishes multi-dimensionality. The proposed model can extract complex features with tomato leaf disease identification accuracy of 96.81\%.

E. Suyawati et~al. \cite{suryawati2018deep} analyze the effect of different depth of CNN architectures (like VGG16, AlexNet, GoogleNet) on a tomato leaf image dataset. The results show that deeper CNN architecture yields better results. P. TM et~al. \cite{tm2018tomato} also propose an approach to detect disease of tomato leaf images using LeNet, a special convolution neural network.

K. Zhang \cite{zhang2018can} uses transfer learning approach to identify tomato leaf disease. Here the CNN model uses AlexNet, GoogleNet, ResNet as the backbone. SGD and Adam are used as optimization methods. Optimal ResNet with SGD shows the highest accuracy of 97.28\%.

\textbf{Research Gap:} 
From the above discussion we see that localization of diseased leaves from images is still a growing research discipline that needs more attention. Also, the majority of the existing detection models are not lightweight enough in terms of computational resource requirement. Besides, large detection networks are prone to overfitting in the case of smaller datasets. Accuracy level of the existing works can also be tried to improve as compared to other domains of deep learning. Hence we undertake the investigation of honing deep learning models for plant disease detection from leaf images. In particular, we investigate Faster-RCNN algorithm's effectiveness and efficiency on tomato leaf images.


\subsection{Background Study}
This subsection highlights some knowledge required to understand the rest of the paper, especially our proposed methods.

\subsubsection{Artificial Neural Network}
\begin{figure}[!h]
\centering
	\includegraphics[height = 4cm, width = 6cm]{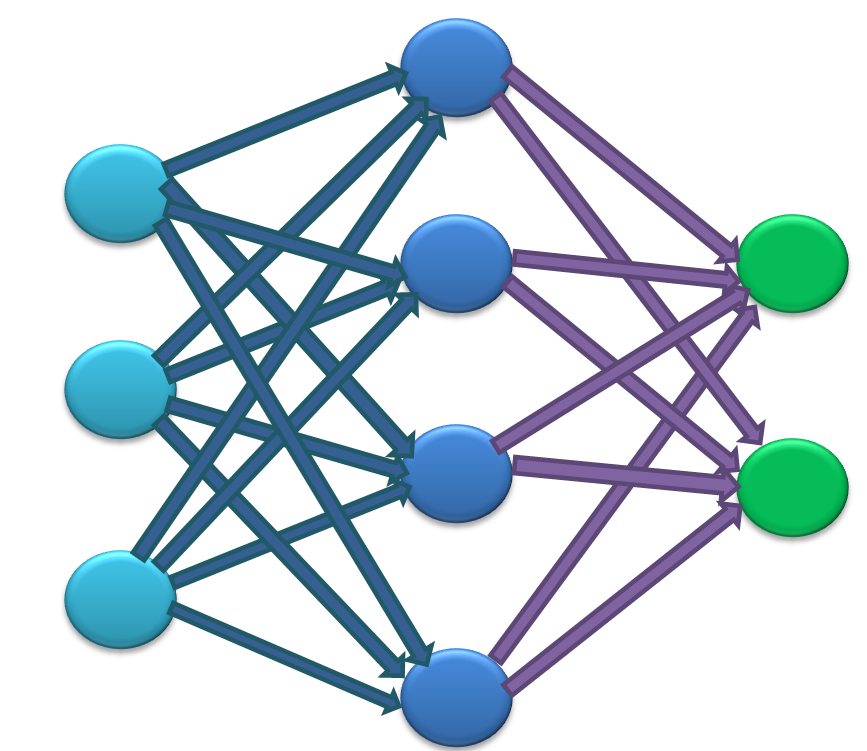}
	\caption{Artificial neural network}
	\label{fig:ANN}
\end{figure}

Artificial Neural Network (ANN) \cite{ann} is one of the most effective machine learning algorithms. It has a layered architecture where at every layer there are several neurons known as perceptrons trying to mimic the decision-making capability of the human brain. The neurons in each layer are fully connected with the previous layer. The input layer receives data, while the output layer delivers the outcome. One or even more hidden layers are placed between the input and output layers. ANN is very effective in real-world applications involving complex features provided enough data are available to it. Figure~\ref{fig:ANN} illustrates a architecture of a simple ANN. In the domain of image processing and computer vision, a 2-dimensional images must be translated to 1-dimensional vectors for ANN which rapidly increases the number of trainable parameters. This requires large amount of storage and processing power which is expensive.
\subsubsection{Convolutional Neural Network}
CNN \cite{cnn} is a special type of deep learning algorithm designed to specially deal with image dataset. 
\begin{figure}[!h]
	\centering
	\includegraphics[height = 5cm, width = 12cm]{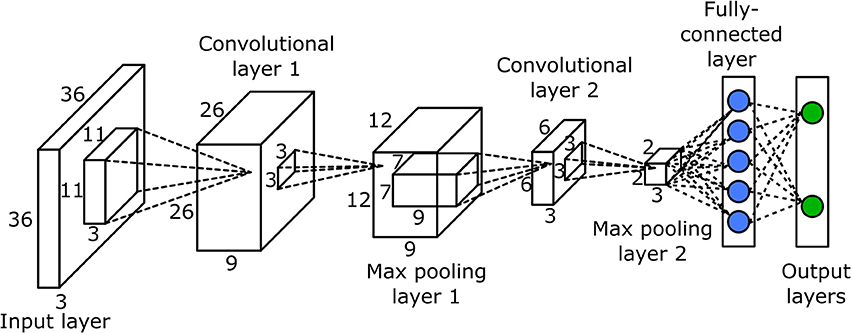}
	\caption{Convolutional neural network \cite{cnn_fig}}
	\label{fig:CNN}
\end{figure}
It takes images as an input matrix of pixel values, assigns weights and biases to different aspects of the objects within the image, and learns to differentiate them automatically without any hand-engineered feature extraction. It adds to the benefits of bare ANN in extracting low-level features like edges, corners and other shapes.

The CNN architecture compresses the entire image into a class scoring vector organized along on the depth plane. Instead of having only fully connected neurons, CNN predominantly has three different types of layers, namely convolutional (ConvNet), Pooling and Fully Connected (Dense) Layers. Among them, convolutional layers are the major components of a CNN network. The architecture of a simple CNN model is shown in Figure~\ref{fig:CNN}.

\subsubsection{Region-Based CNN}
\begin{figure}[!h]
	\centering
	\includegraphics[height = 4cm, width = 14cm]{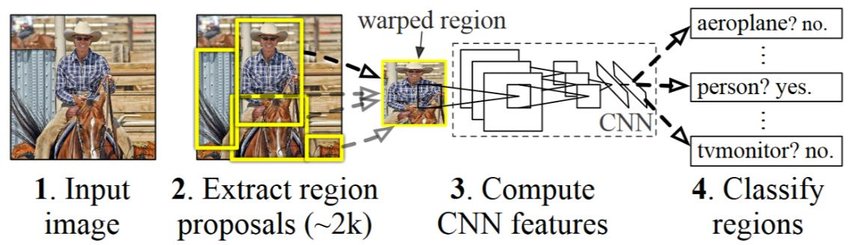}
	\caption{Region-Based CNN \cite{rcnn}}
	\label{fig:RCNN}
\end{figure}

In 2014, Girshick et~al. \cite{rcnn} propose a CNN-based object detection algorithm named Region-based CNN (RCNN) to overcome the difficulty of selecting the massive amount of regions from an image. Here, selective search is used for extracting only 2000 regions of interest from the image which are called ``region proposals''. The region proposals are fed into the CNN network  as input for getting the feature vectors. CNN works as the feature extractor network. Finally, the output layer of the feature extractor is fed into an SVM \cite{svm} classifier for detecting the presence of an object within the proposed region. Figure~\ref{fig:RCNN} illustrates the overall organization of the RCNN model.

\subsubsection{Fast RCNN}
\begin{figure}[!h]
	\centering
	\includegraphics[height = 4cm, width = 11cm]{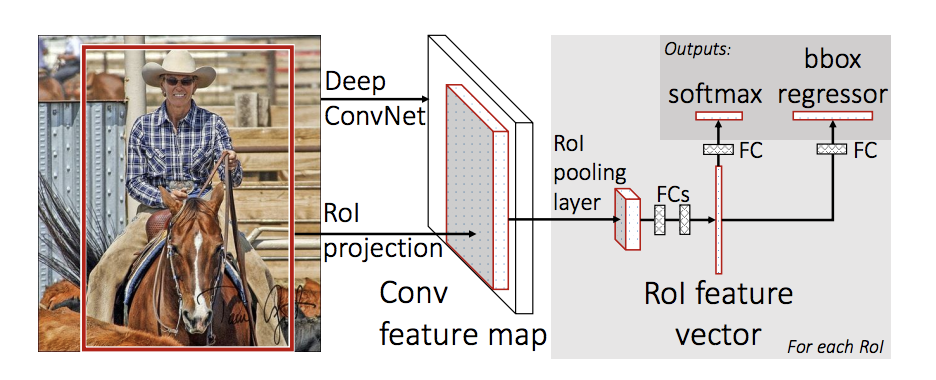}
	\caption{Fast RCNN (\cite{a20})}
	\label{fig:Fast-RCNN}
\end{figure}

RCNN suffers from high computational time requirement. To mitigate this problem, better version of RCNN, named Fast RCNN \cite{a20} is introduced in 2015 by the same authors. Fast RCNN is requires less time than RCNN as it avoids feeding 2000 region proposals to the CNN. Rather, input images are directly fed to the CNN to generate feature maps. The region of proposals are identified from the feature map and then warped into squares. They are reshaped into a fixed size so that they can be fed into a fully connected layer, by using an RoI (Region of Interest) pooling layer. A softmax layer is then used to predict the proposed region's class and the offset values for the bounding box. The overall working procedure of Fast RCNN is shown in Figure~\ref{fig:Fast-RCNN}. However, the bottleneck of Fast RCNN is the region proposal generation which is performed by selective search.

\subsubsection{Faster RCNN}
\begin{figure}[!h]
	\centering
	\includegraphics[height = 9cm, width = 15cm]{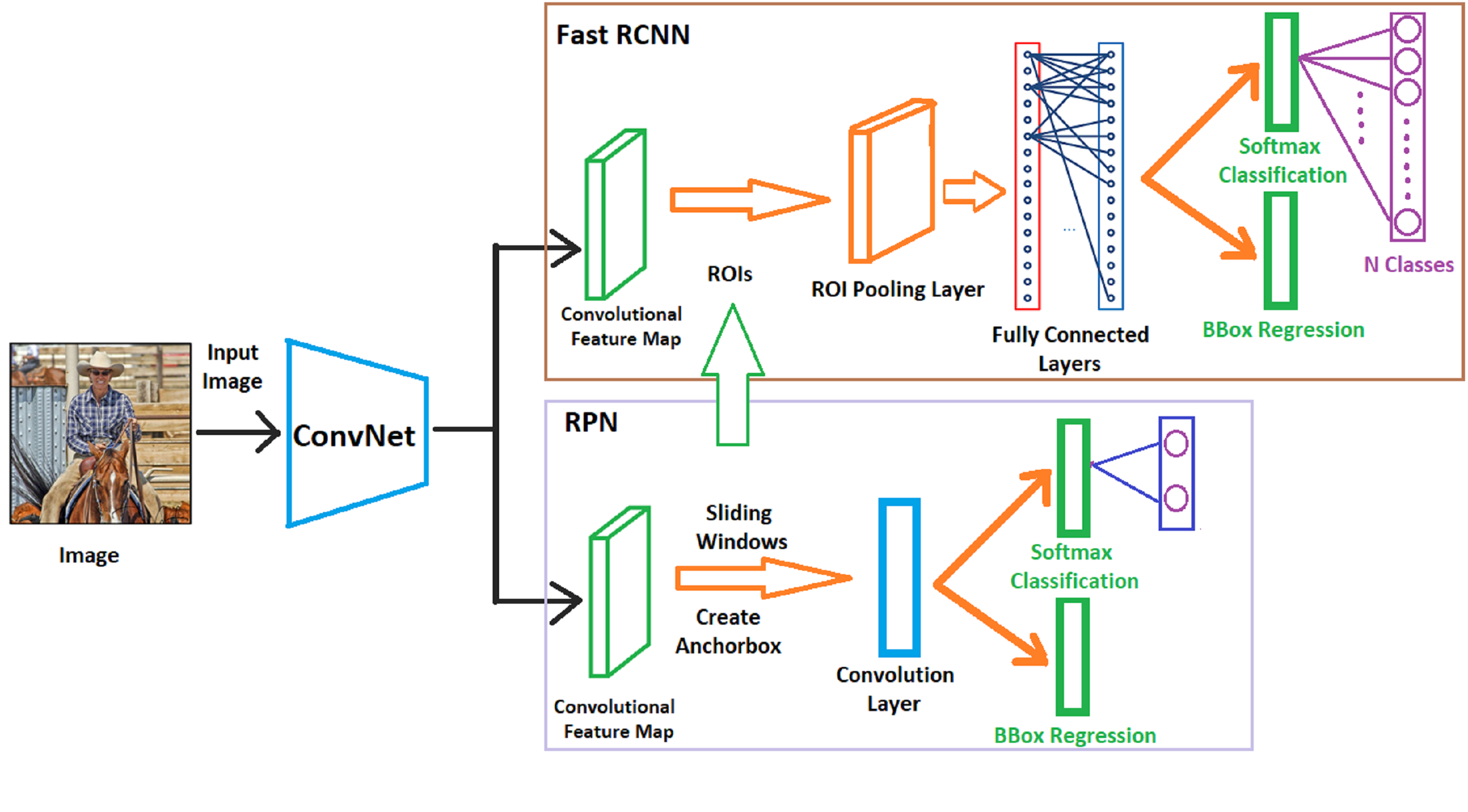}
	\caption{Faster RCNN \cite{a21}}
	\label{fig:Faster-RCNN}
\end{figure}

Faster RCNN is the final version of this state-of-the-art method \cite{a21}. It is a single-stage architecture that is able to be trained end-to-end. In this architecture a region proposal network (RPN) is used which is responsible for generating region proposals. Compared to traditional algorithms like selective search, this method takes less computational time because it uses the GPU computation. To extract a fixed-length feature vector from each region proposal, Faster RCNN uses the ROI Pooling layer.  Figure~\ref{fig:Faster-RCNN} visualizes the architecture of Faster RCNN.\footnote{For PASCAL VOC 2007 dataset, Fast RCNN and Faster RCNN obtain  66.9 and 70.4 mean average precision respectively.}

RPN investigates bounding boxes and class scores at each location concurrently. RPN is trained end-to-end for generating high-quality region proposal generation. Anchor boxes at various scales and aspect ratios are used by RPN as references. The responsibility of RPN is to effectively predict Region Proposals with a wide range of scales and aspect ratios. It  slides an window over the feature map to generate ``proposals'' for the locations where the object may encounter. The RPN Classifier calculates the likelihood of a proposal to have the target object. The overview of the RPN is shown on the lower part of Figure~\ref{fig:Faster-RCNN}.

\section{Proposed Approach}
\label{sec:proposed}
We propose an object detection model named Improved RCNN for detecting plant leaf disease. The whole architecture is a collection of three building blocks which are: (a) Feature Map Generator Network, (b) Region of Interest (RoI) Generator Network, and (c) Detector Network. These modules are described below, which is followed by a discussion on training the proposed model.

\subsection{Architecture of Improved RCNN}

\begin{figure}[!h]
	\centering
	\includegraphics[height = 9cm, width = 16cm]{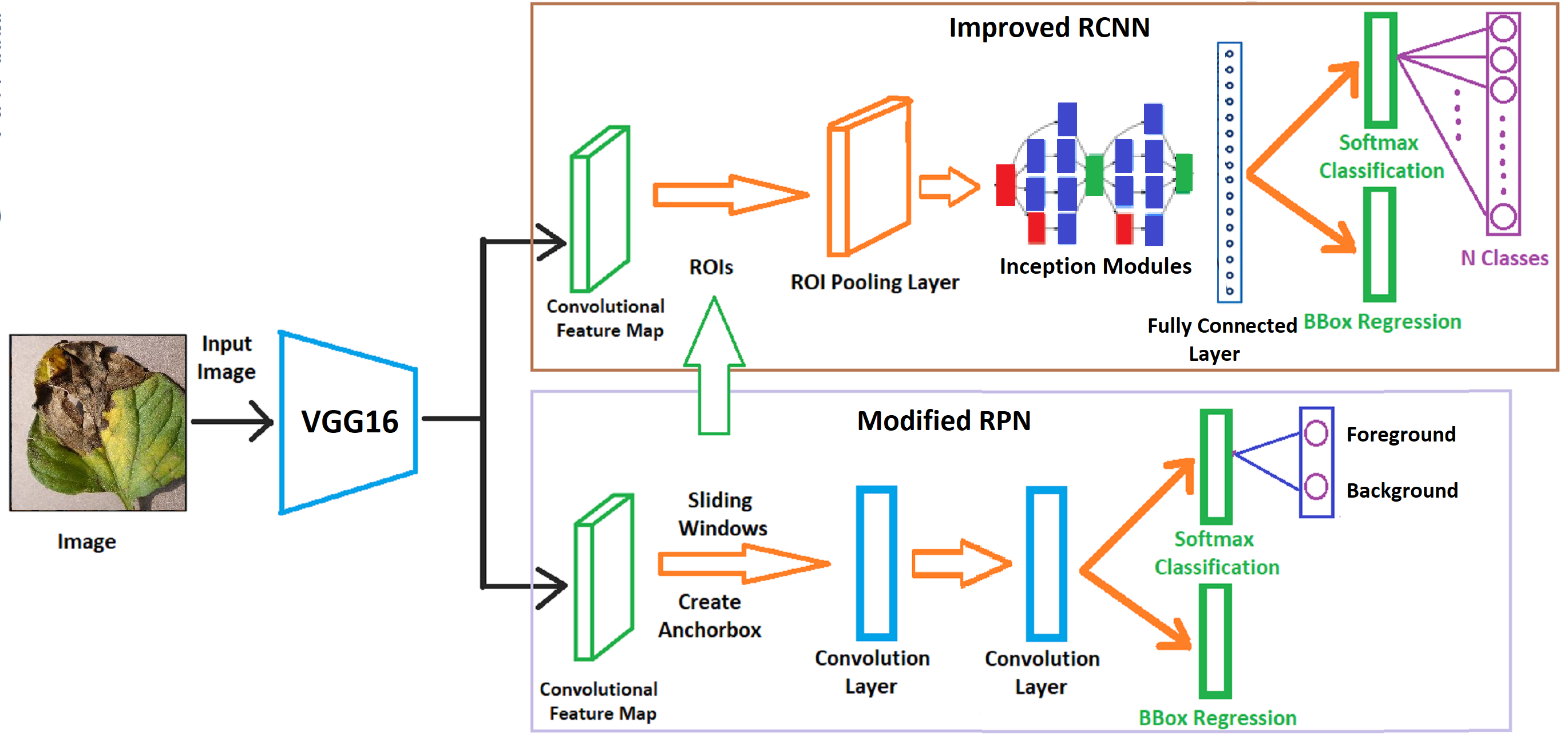}
	\caption{Proposed Framework}
	\label{fig:Proposed Framework}
\end{figure}
The proposed object detection model for tomato leaf disease identification based on Improved RCNN is depicted in Figure~\ref{fig:Proposed Framework}. We use pre-trained VGG16 as the backbone network to generate feature maps. For proposing the region of interest, we use a variant of the RPN. The image of tomato leaf is fed into the base network for feature generation.The RPN and the detector network both exploit this feature vector. IoU optimization is used for improving region proposals. The architecture of our proposed framework is briefly described in the following subsections.

\subsubsection{Feature Map Generator Network: VGG16}
Improved RCNN requires a pre-trained model for feature extraction. We chose VGG16 as it is widely used for transfer learning. Initially, VGG16 has 12 convolution layers and 5 maxpooling layers. Instead of using a variety of filters, padding, and stride, VGG16 used 3 × 3 filters for convolutional every layer and 2 × 2 filters for maxpooling. The entire architecture has a bunch of convolution and maxpool layers and ends with two fully connected layers each having 4096 nodes.  With a large number of trainable parameters (138M) the training time of VGG16 is very high. However, we remove the fully connected layers of the network reducing the number of parameters to only 14.7M. The last maxpooling layer is replaced by the ROI pooling before going into the inception module of Improved RCNN later. VGG16 network takes 224x224 sized images as the input. We modify the input image size to 256 × 256 to fit our dataset. For each image of size 256 × 256  × 3 a feature maps of size 512 × 16 × 16 generated from the last layer of the pretrained VGG16 network. It is used as the input for the modified RPN as well as the Improved RCNN classifier Network.

\subsubsection{RoI Generator Network: Modified RPN}
The Modified RPN is the backbone of Improved RCNN. RPN is a fully convolutional network responsible for finding interesting regions. Modified RPN contains two consecutive fully convolutional layers each having 512 filters of size 3 × 3. Anchor boxes are being used as references at various scales and aspect ratios. The responsibility of RPN is to effectively predict Region Proposals from the feature map. A Sliding window  slides over the feature map to generate “proposals” for the interesting region. This fully convolutional Neural Network has two output branches. One branch is responsible for generating a set of rectangular object or region proposals and another one is responsible for generating a score that determines whether the region is foreground or background. For each spot on the feature map, softmax classifier results in 9 values for $k = 9$. The RPN Classifier calculates the likelihood of a proposal to contain the object of interest. The RPN Regressor is responsible for predicting the offset of the coordinates of the proposal. For 9 anchorboxes, the number of coordinate values returned by the regressor node is $9 × 4 = 36 $ for each point on the feature map. For a feature map of size $W$ × $H$, the number of anchors is $W$ × $H$ × $k$.

\subsubsection{Classifer Network}
The last maxpool layer of VGG16 is swapped with an ROI pooling layer to prepare the feature map of any size of to a fixed one. Region of Interest (ROI) is a rectangle-shaped window which is defined by $(x,y,w,h)$ where $(x,y)$ is the coordinate of the top left corner and w, h are width and height respectively. ROI pooling layer uses maxpooling for converting features inside the ROIs into fixed-sized (H×W) feature maps. Firstly, the $(h × w)$ sized ROI is divided into $(H×W)$ grid where each grid size is $(h/H) × (w/W)$. Then, maxpooling is performed to each grid independently to get the output ROI map of size $(H×W)$. After ROI pooling the feature maps are ready to be fed into the first fully connected layer of the classifier network.

\begin{figure}[!h]
	\centering
	\includegraphics[height = 5cm, width = 9cm]{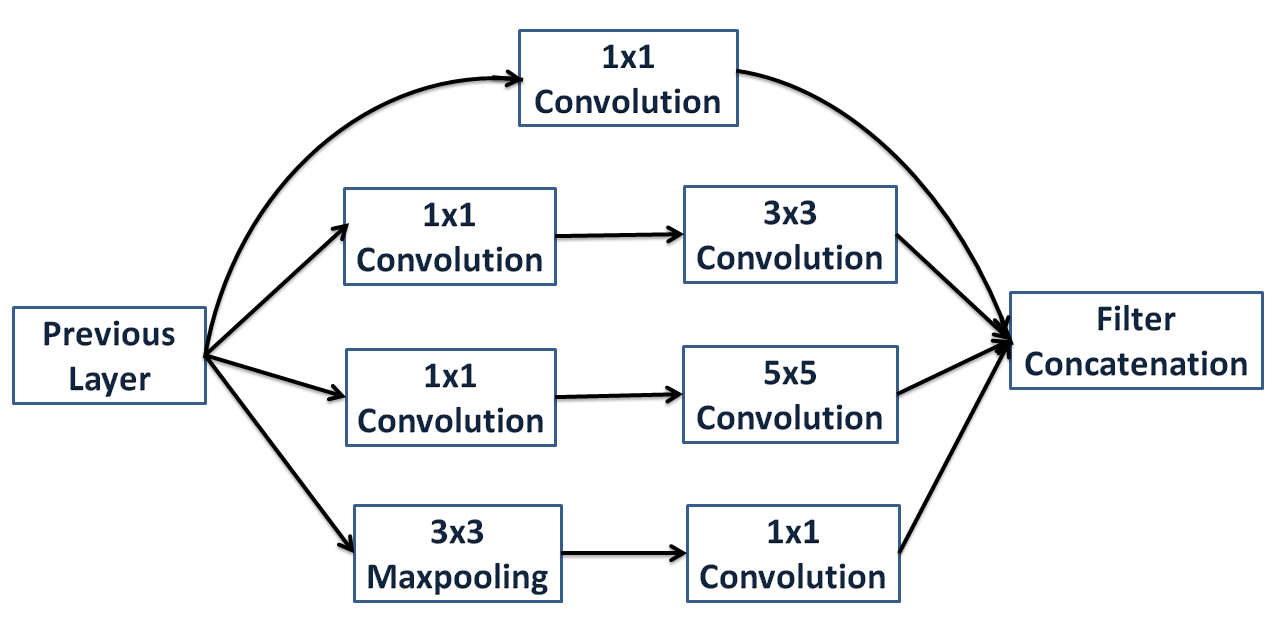}
	\caption{Inception Module}
	\label{fig:Inception Module}
\end{figure}

The output feature maps of the ROI pooling layer along with a number of ROIs and number of output classes are fed into the classifier network i.e. Improved RCNN. In the Fast RCNN, there are two fully connected (FC) layers with 50\% dropout where each FC layer has 4098 neurons. These two fully connected layers are the most costly part of the Faster RCNN network in terms of the number of parameters and thereby this huge network used a substantial amount of computing resources and is also prone to overfitting for smaller datasets. The most effective way to improve feature extraction capability without increasing the depth of the network is to introduce parallel layers. Inception modules have parallel layers with different sized convolution kernels and finally concatenate their results. This results in an enhancement of the adaptability of the network to different scales (\cite{aa1}) and also helps to extract good features from low quality images also. Inception modules decrease the number of parameters while maintaining the range of perspective fields also. It consists of three different sized convolution filters (1 × 1, 3 × 3, and 5 × 5) and a 3 × 3 maxpooling. The outputs of these parallel layers are finally concatenated to get the resulting output. In the inception module, the convolution operations are performed on the same level. We replace one fully connected layer with two consecutive inception modules. To reduce the computational expense more, an extra 1x1 convolution is added before the 3 × 3 and 5 × 5 filters and after 3 × 3 maxpooling layers. The structure of an Inception module used in our detector network is illustrated in Figure~\ref{fig:Inception Module}. The second Inception module is followed by a fully connected layer with 1024 neurons. 50\% dropout is enforced to reduce overfitting. The FC layer is branched into two output nodes. One of them is responsible for calculating the softmax probability estimations for $N+1$ distinct classes. The another sibling is responsible for calculating  the bounding box offset values $(x1,y1,x2,y2)$ for the corresponding $N$ classes.

\subsection{Training and Optimization of Improved RCNN}
Training and optimization of modified RPN can be done end to end by using backpropagation and SGD (stochastic gradient descent). The Improved RCNN detector network can also be optimized separately. But training these two parts separately will result in modifying the shared convolutional layers differently. Another alternative can be considering the whole framework as a single network and trying to optimize it using backpropagation jointly. But this way is also infeasible because the training phase of Improved RCNN is dependent on fixed object proposals. So changing the proposals simultaneously may hamper the convergence of the network drastically. A step-by-step optimization \cite{a21} is being used to overcome the difficulty of optimizing the network whilst taking advantage of shared convolutional layers, as shown below.

\begin{itemize}
\item The modified RPN network is trained for region proposal task with VGG16 pre-trained model. The modified RPN is finetuned end to end here.
\item The Region Proposals from modified RPN are used to train the Improved RCNN detection network
\item The Improved RCNN is used to initialize the RPN. Only the fully convolutional layers exclusive to the RPN network are fine-tuned. The two networks start to share the convolution layers of VGGNet here but the weights are fixed. Finally the fully connected layers of the Detector Network, i.e. Improved RCNN, are finetuned keeping the weights of shared layers fixed.
\end{itemize}
\section{Performance Evaluation}
\label{sec:result}
In this section, we introduce the experimental settings, describe the dataset~along with preprocessing steps, and discuss the results.

\subsection{Experimental Settings}
To evaluate our proposed model, a benchmark publicly available dataset is used after necessary preprpossing. The statistics of the dataset and preprocessing operations are briefly described in Section~\ref{sec:dataset}. The experiments are carried out using a GPU environment whose various settings are mentioned in Section~\ref{sec:implementation details}.

\subsubsection{Dataset Description}
\label{sec:dataset}

Plantvillage dataset \cite{a18}\footnote{\url{https://www.kaggle.com/datasets/abdallahalidev/plantvillage-dataset}} is an open access repository with 54303 plant leaf images categorized into 38 classes. From this dataset, we use only the tomato leaf images mainly due to the extensive time requirement of image annotation process. 

A total of 9343 images of diseased tomato leaves are separated. 
The dataset is divided into 70:20:10 for training, validation, and testing respectively. Nine most common tomato  leaf diseases used for our experiment are Early Blight, Bacterial Spot, Target Spot, Late Blight, Septoria Leaf Spot, Leaf Mold, Yellow Leaf Curl Virus, Two Spotted Spider Mite, and Mosaic Virus. The statistics of the dataset are given in Table~\ref{tab:data}.
	\begin{table}[]
		\centering
		\caption{Summary of Dataset}
		\label{tab:data}
		\begin{tabular}{p{0.2\textwidth}*{3}{>{\centering\arraybackslash}p{0.2\textwidth}}}
			\hline
			\textbf{Class}                  & \textbf{Train} & \textbf{Validation} & \textbf{Test} \\ \hline
			\textbf{Bacterial spot}         & 700            & 200                 & 100           \\
			\textbf{Early blight}           & 700            & 200                 & 100           \\
			\textbf{Healthy}                & 691            & 197                 & 98            \\
			\textbf{Late   blight}          & 678            & 194                 & 97            \\
			\textbf{Leaf Mold}              & 666            & 190                 & 96            \\
			\textbf{Septoria leaf spot}     & 700            & 200                 & 100           \\
			\textbf{Spider mite}            & 700            & 200                 & 100           \\
			\textbf{Target Spot}            & 700            & 200                 & 100           \\
			\textbf{Mosaic virus}           & 261            & 75                  & 37            \\
			\textbf{Yellow Leaf Curl Virus} & 744            & 213                 & 106           \\ \hline
			\textbf{Total}                  & 6540           & 1869                & 934           \\ \hline
		\end{tabular}
	\end{table}
\begin{figure}[!h]
	\centering
	\includegraphics[height = 8cm, width = 15cm]{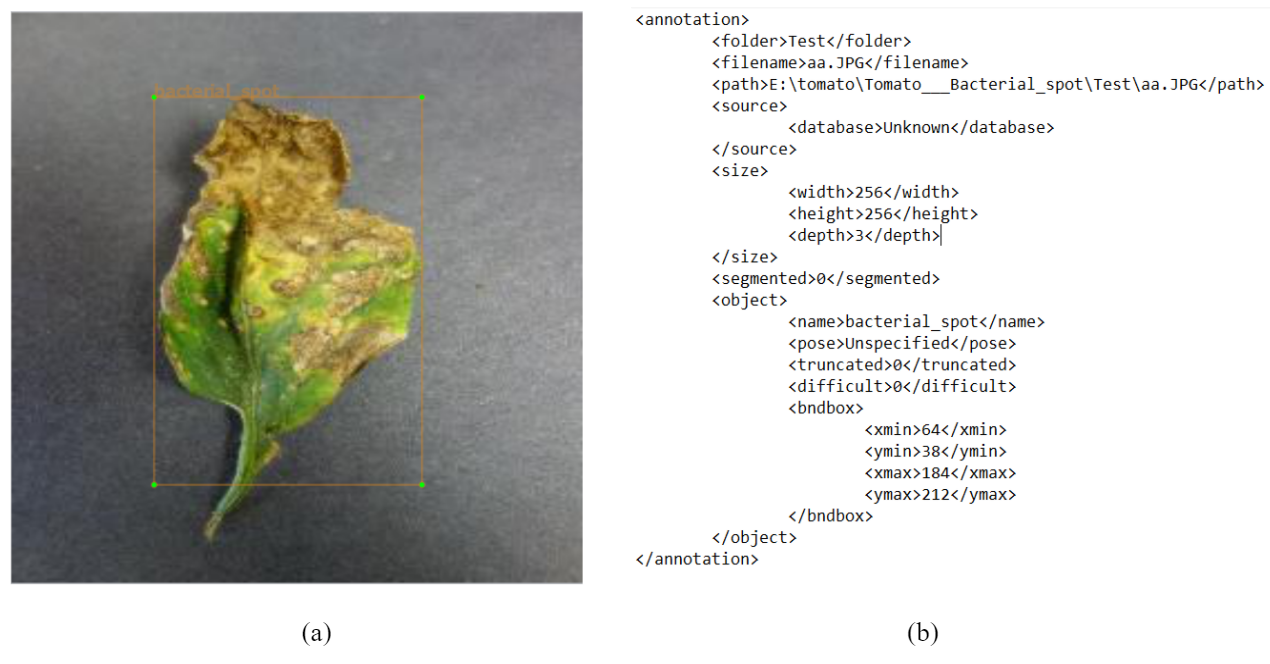}
	\caption{Annotation of Tomato leaf dataset (a) Annotated image with bacterial spot disease, (b) XML document for corresponding image.}
	\label{fig:annotation}
\end{figure}
Each image is resized to 256x256 pixels. Image annotation is then performed using a python tool named LabelImg\footnote{\url{https://github.com/tzutalin/labelImg.git}}. 
The objective of image annotation is to label the exact position of diseased leaves in the images. All the leaf images are annotated using a python program. At the completion of the annotation, for every image a corresponding XML file is generated. The XML file includes the location of the diseased leaf as well as the coordinates of the upper left and lower right corners of the subject area. Annotation of a sample image with bacterial spot disease is shown in Figure~\ref{fig:annotation}.

\subsubsection{Implementation Details}
\label{sec:implementation details}
This section briefs about the implementation details of the experiments. The  models are trained for 300 epochs with a learning rate of 0.00001. We use one of the most popular optimizing algorithms named Adam optimizer for the learning phase. Anchor box scales  are $60^2$, $120^2$, and $240^2$ considering 4 ROIs per pixel position.  The hyperparameter settings for all the investigated models are mentioned in Table~\ref{tab:hyperparam}. 

\begin{table}[]
\centering
\caption{Hyperparameter Settings used for all the investigated models}
\label{tab:hyperparam}
\begin{tabular}{p{0.2\textwidth}*{3}{>{\centering\arraybackslash}p{0.2\textwidth}}}
\hline
Hyperparameter & Value     \\ \hline
Image Size     & 256x256x3 \\
Epochs         & 300       \\
Dropout Rate   & 0.5       \\
Learning Rate  & 0.00001   \\
Optimizer      & Adam      \\ \hline
\end{tabular}
\end{table}

The experiments are conducted on a Windows machine with an Intel® Core™ i5-10300H CPU @ 2.50GHz accelerated by an NVIDIA GeForce RTX 2060 GPU. NVIDIA GeForce RTX 2060 GPU has 1920 CUDA Cores and 6GB memory. The core frequency is up to 1365 MHz. The models are implemented in the Tensorflow\footnote{\url{https://www.tensorflow.org/}} framework. All the hardware and software specifications are listed in Table~\ref{tab:hardware}.

\begin{table}[]
			\centering
			\caption{Experimental Environment}
			\label{tab:hardware}
	\begin{tabular}{p{0.2\textwidth}*{3}{>{\centering\arraybackslash}p{0.5\textwidth}}}
		\hline
		\textbf{Equipments} & \textbf{Specifications}                \\ \hline
		\textbf{System}     & Windows 10                             \\
		\textbf{Framework}  & Tensorflow 2                           \\
		\textbf{Language}   & Python 3.7                             \\
		\textbf{CPU}        & Intel® Core™ i5-10300H   CPU @ 2.50GHz \\
		\textbf{GPU}        & NVIDIA GeForce RTX 2060   6GB          \\
		\textbf{Memory}     & 12GB                                   \\ \hline
	\end{tabular}
\end{table}

For image processing and especially for object detection, precision and recall are two heavily used metrics. Precision is the proportion of relevant classes vs. retrieved classes. Recall is the proportion of the retrieved relevant instances. F2 score is another metric that calculates an average of precision and recall providing recall the higher weight. The equations for retrieving precision, recall, and F2 scores are as follows.
\begin{equation}\label{1}
\centering
precision=\frac{TP}{TP+FP}
\end{equation}

\begin{equation}\label{2}
\centering
recall=\frac{TP}{TP+FN}
\end{equation}

\begin{equation}\label{3}
\centering
F2=\frac{TP}{TP+0.2FP+0.8FN}
\end{equation}

Here,\\
$TP$= \emph{True Positive (Number of correctly classified positive instances)}\\
$FP$= \emph{False Positive (Number of correctly classified negative instances)}\\
$FN$= \emph{False Negative (Number of wrongly classified positive instances)}

\subsection{Result Analysis}
\begin{figure}[!h]
	\centering
	\includegraphics[height = 5cm, width = 15cm]{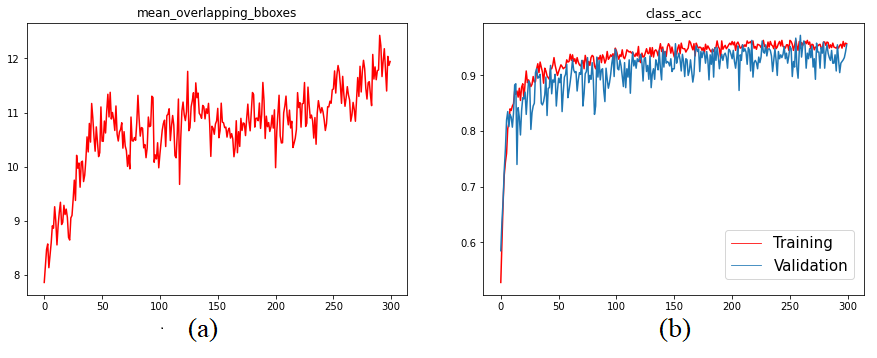}
	\caption{: (a) Mean overlapping Bounding Boxes, and (b) Classification Accuracy over the epochs for Improved RCNN}
	\label{fig:(a) Mean overlapping Bounding Boxes,  (b) Classification Accuracy over the epochs.}
\end{figure}


\begin{figure}[!h]
	\centering
	\includegraphics[height = 5cm, width = 15cm]{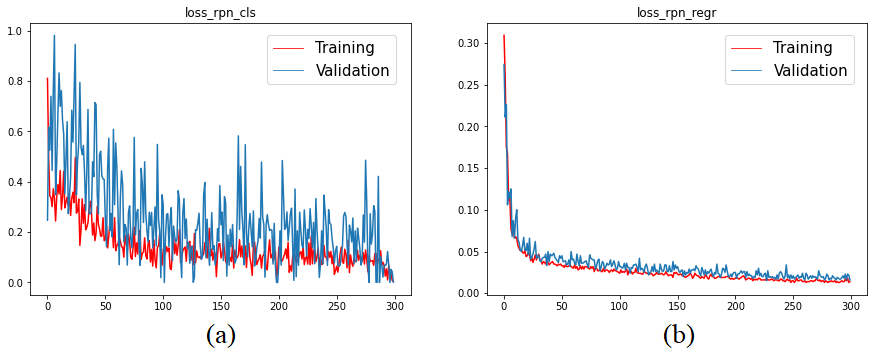}
	\caption{(a) Loss curve for modified RPN classifier, and (b) loss curve for modified RPN regression for Improved RCNN}
	\label{fig: (a) Loss curve for RPN class, (b) Loss Curve for RPN Regression}
\end{figure}


While considering 0.7 Intersection over Union (IoU) the number of mean overlapping bounding boxes over the period is shown in Figure~\ref{fig:(a) Mean overlapping Bounding Boxes,  (b) Classification Accuracy over the epochs.}a. Figure~\ref{fig:(a) Mean overlapping Bounding Boxes,  (b) Classification Accuracy over the epochs.}b visualizes the improvement curve of classification accuracy for 300 epochs with a learning rate of 0.00001. 
The classification accuracy after 300 epochs is found to be 96.4\%.

The average cross-entropy of each epoch during the training process is represented by loss. A lower loss value indicates a better classification performance of the model. As there are 4 output nodes in the architecture, 4 different losses are considered for each of them.
For the RPN network, there are two losses for classification and regression. The final loss for modified RPN classification is found to be 0.004. For modified RPN regression it is slightly higher than 0.013. Figure~\ref{fig: (a) Loss curve for RPN class, (b) Loss Curve for RPN Regression} shows the loss curves for modified RPN classification and regression.

\begin{figure}[!h]
	\centering
	\includegraphics[height = 5cm, width = 15cm]{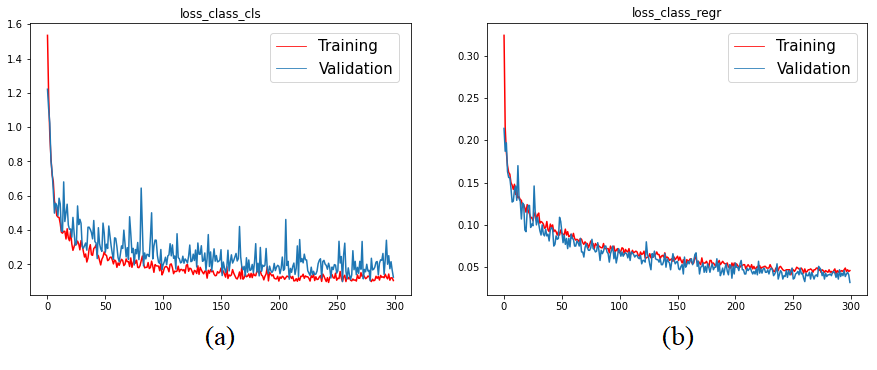}
	\caption{(a) Loss curve for classification, and (b) loss curve for regression  for Improved RCNN}
	\label{fig:(a) Loss curve for classification, (b) Loss Curve for Classification Regression}
\end{figure}

For the Improved RCNN classifier module, there are also two losses: one for classification and another for regression. The loss for detector classifier and regressor are 0.10 and 0.04 respectively. Figure~\ref{fig:(a) Loss curve for classification, (b) Loss Curve for Classification Regression} illustrates the loss curves for the Improved RCNN detector module. The total loss for the network is 0.16.

\begin{figure}[!h]
	\centering
	\includegraphics[height = 6cm, width = 8cm]{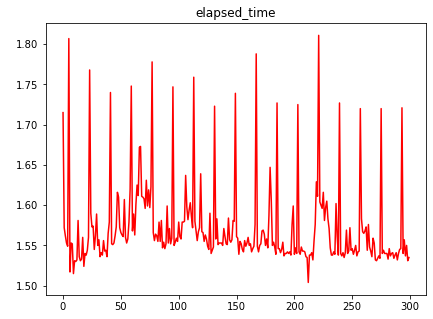}
	\caption{Training time per epochs for Improved RCNN}
	\label{fig:Training time per epochs.}
\end{figure}

The elapsed time for each training epoch is around $1.65 ± 0.1$ minutes. Figure~\ref{fig:Training time per epochs.} visualizes the training time per epoch. However, the testing time for each image is 1.67 seconds. The mean average precision for the Improved RCNN is 89.5\%.

\begin{table}[]
			\centering
			\caption{Classification accuracy VS bounding box threshold of Improved RCNN model}
			\label{tab:Classification accuracy VS bounding box threshold of Improved RCNN model}
	\begin{tabular*}{\textwidth}{c @{\extracolsep{\fill}} ccccccccc}
		\hline
		\textbf{Class vs Bbox threshold} & \textbf{0.1}           & \textbf{0.2} & \textbf{0.3}           & \textbf{0.4}           & \textbf{0.5}           & \textbf{0.6}           & \textbf{0.7}           & \textbf{0.8}           & \textbf{0.9} \\ \hline
		\textbf{Early Blight}            & \textit{\textbf{0.97}} & 0.97         & 0.97                   & 0.97                   & 0.97                   & 0.97                   & 0.96                   & 0.94                   & 0.90         \\
		\textbf{Spider Mites}            & 0.96                   & 0.96         & \textit{\textbf{0.97}} & 0.97                   & 0.96                   & 0.96                   & 0.96                   & 0.96                   & 0.94         \\
		\textbf{Healthy}                 & 0.98                   & 0.98         & 0.98                   & \textit{\textbf{0.99}} & 0.99                   & 0.99                   & 0.99                   & 0.99                   & 0.99         \\
		\textbf{Late Blight}             & \textit{\textbf{0.96}} & 0.96         & 0.96                   & 0.95                   & 0.94                   & 0.94                   & 0.95                   & 0.93                   & 0.93         \\
		\textbf{Target Spot}             & 0.96                   & 0.96         & 0.96                   & 0.96                   & \textit{\textbf{0.97}} & 0.95                   & 0.95                   & 0.95                   & 0.95         \\
		\textbf{Bacterial Spot}          & 0.97                   & 0.97         & 0.97                   & 0.97                   & \textit{\textbf{0.98}} & 0.98                   & 0.98                   & 0.98                   & 0.97         \\
		\textbf{Leaf Mold}               & 0.99                   & 0.99         & 0.99                   & 0.99                   & 0.99                   & 0.99                   & 0.99                   & \textit{\textbf{1.00}} & 1.00         \\
		\textbf{Yellow Leaf Curl Virus}  & 0.99                   & 0.99         & 0.99                   & 0.99                   & 0.99                   & 0.99                   & \textit{\textbf{1.00}} & 1.00                   & 1.00         \\
		\textbf{Septoria Leaf Spot}      & 0.98                   & 0.98         & 0.98                   & 0.98                   & 0.98                   & \textit{\textbf{0.99}} & 0.98                   & 0.97                   & 0.97         \\
		\textbf{Mosaic Virus}            & \textit{\textbf{0.99}} & 0.99         & 0.99                   & 0.99                   & 0.99                   & 0.99                   & 0.99                   & 0.99                   & 0.99         \\ \hline
		\textbf{Overall}                   & 0.97                   & 0.97         & 0.97                   & 0.97                   & 0.97                   & \textit{\textbf{0.98}} & 0.97                   & 0.97                   & 0.96         \\ \hline
	\end{tabular*}
\end{table}

For improving the detection accuracy of the object classes inside the bounding box, optimization is done. For this purpose, the model is evaluated using different bounding box thresholds from 0.1 to 0.9 for 3738 validation data of 10 different classes. The initial threshold is 0.1 for all the classes and the classification accuracy for 0.1 is stored for further comparison. For the next thresholds whenever the classification accuracy increased for a class the best threshold is updated. The results are listed in Table~\ref{tab:Classification accuracy VS bounding box threshold of Improved RCNN model}. The final best thresholds for each class are highlighted in the table.

\begin{table}[]
		\centering
		\caption{Evaluation Result of Improved RCNN}
		\label{tab:Evaluation Result of Improved RCNN}
	\begin{tabular*}{\textwidth}{c @{\extracolsep{\fill}} cccccc}
		\hline
		\textbf{Class Name}             & \textbf{True Positive}  & \textbf{False Positive} & \textbf{False Negative} & \textbf{F2 Sores} & \textbf{Precision} & \textbf{Recall} \\ \hline
		\textbf{Early Blight}           & 100           & 8          & 0           & 0.9821              & 0.9258               & 0.9999            \\
		\textbf{Spider Mites}           & 98           & 6           & 4           & 0.9549              & 0.9422               & 0.9607            \\
		\textbf{Healthy}                & 93           & 1           & 1           & 0.9873              & 0.9893               & 0.9893            \\
		\textbf{Late Blight}            & 86           & 1           & 11          & 0.9033              & 0.9884               & 0.8865            \\
		\textbf{Target Spot}            & 95           & 7           & 1           & 0.9752              & 0.9313               & 0.9895            \\
		\textbf{Bacterial Spot}         & 104          & 4           & 0           & 0.9902              & 0.9629               & 0.9999            \\
		\textbf{Leaf Mold}              & 94           & 0           & 0           & 0.9979              & 0.9999               & 0.9999            \\
		\textbf{Yellow Leaf Curl Virus} & 106          & 0           & 0           & 0.9979              & 0.9999               & 0.9999            \\
		\textbf{Septoria Leaf Spot}     & 102          & 6           & 1           & 0.9786              & 0.9444               & 0.9902            \\
		\textbf{Mosaic Virus}           & 37           & 2           & 0           & 0.9870              & 0.9485               & 0.9997            \\ \hline
		\textbf{Overall}                & \textbf{915} & \textbf{35} & \textbf{18} & \textbf{0.9771}     & \textbf{0.9631}      & \textbf{0.9807}   \\ \hline
	\end{tabular*}
\end{table}

After the optimization procedure, the best thresholds for each class are stored for further usage. Then, the model is evaluated using the optimized thresholds for each class. The detection performance of the model is evaluated in terms of Mean Average Precision (mAP) which is an effective metric of accuracy for object detection tasks and closely related to precision, recall and F2 Score. 

\begin{figure}[!h]
	\centering
	\includegraphics[height = 7cm, width = 7cm]{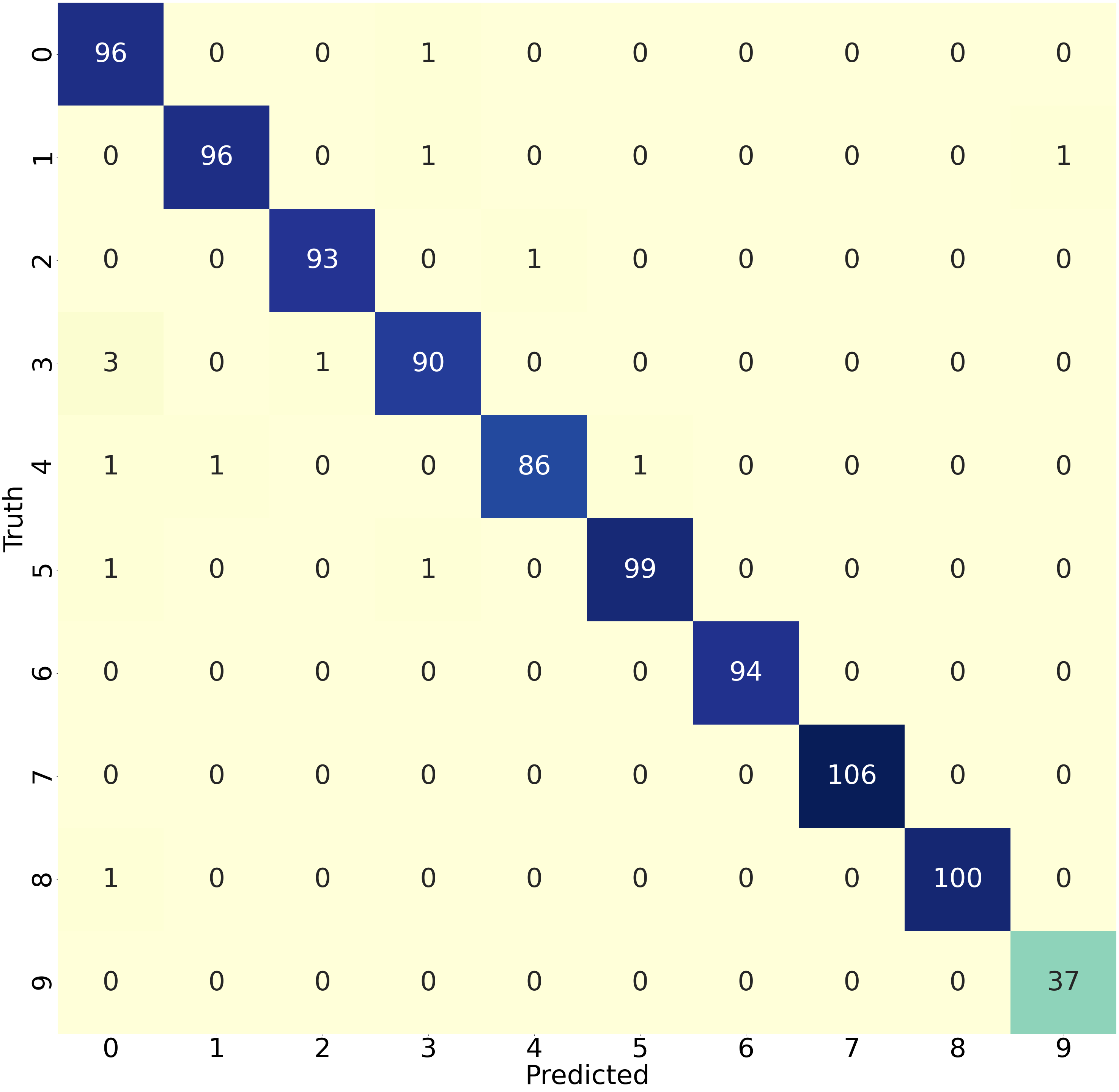}
	\caption{Confusion matrix for Improved RCNN}
	\label{fig:Confusion matrix}
\end{figure}

\begin{figure}[!h]
	\centering
	\includegraphics[height = 6cm, width = 14cm]{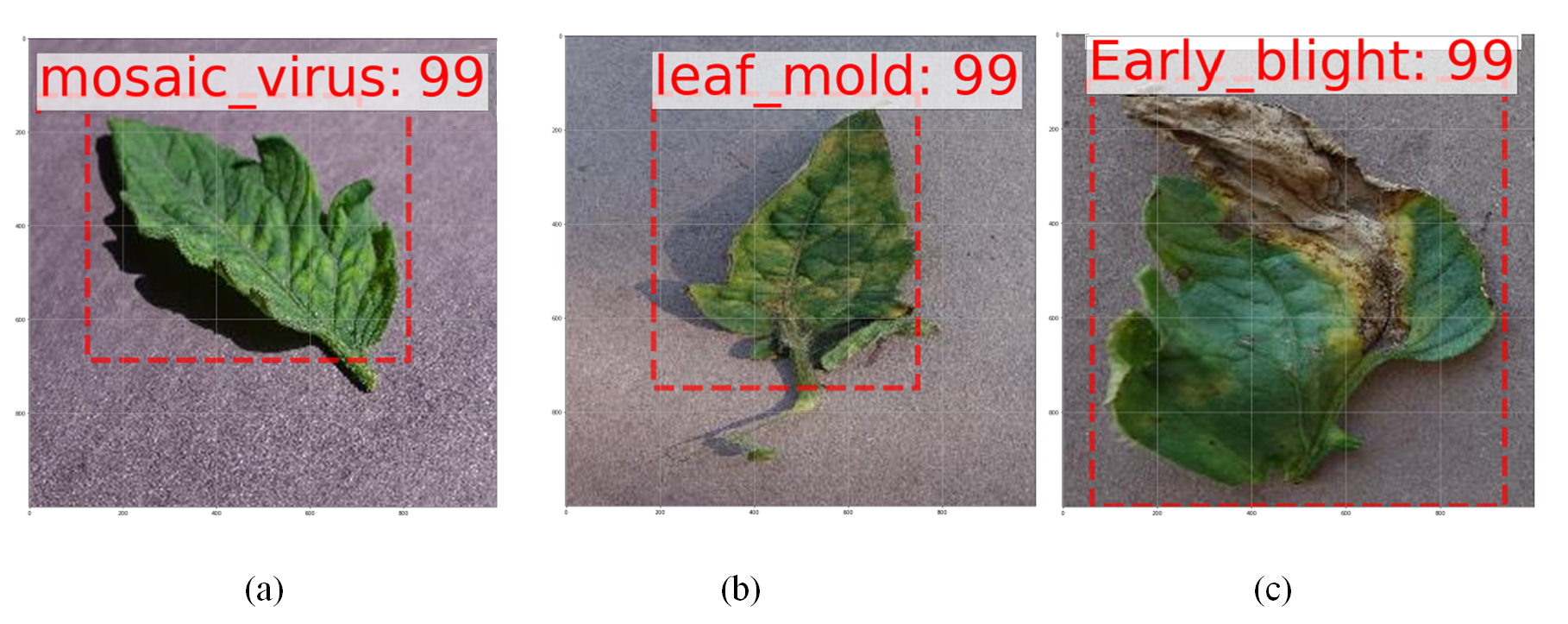}
	\caption{Detection performance of proposed Improved RCNN Architecture for test data from PlantVillage dataset}
	\label{fig:Detection Result of proposed Improved RCNN architecture for test data from PlantVillage}
\end{figure}

The confusion matrix in Figure~\ref{fig:Confusion matrix} visualizes the evaluation performance of our proposed algorithm. Here $X$ and $Y$ axes represent predicted values and the ground truth values for the supervised annotation respectively. The labels 0 to 9 represent Early Blight, Two Spotted Spider Mites, Healthy, Late Blight, Target Spot, Bacterial Spot, Leaf Mold, Yellow Leaf Curl Virus, Septoria Leaf Spot, Mosaic Virus, and no disease respectively. From the figure, it is visible that only a few images are misclassified.

Figure~\ref{fig:Detection Result of proposed Improved RCNN architecture for test data from PlantVillage} illustrates the resulting bounding boxes, class labels, and detection probability for some sample test images. For all the test images the detection probability is between 97\% - 99.997\% which is impressive.

We employ three pre-trained backbone networks, namely (i) VGG16, (ii) VGG19, and (iii) MobileNet, for feature extraction to test (i) the regular Fast RCNN detector with 4096 neurons in each fully-connected (FC) layer, (ii) Fast RCNN with 1024 neurons in each FC layer, and (iii) Inception based Improved RCNN detector. The total number of parameters for each of them is mentioned in Table~\ref{tab:Total Number of Parameters For each Models}. From Table~\ref{tab:Total Number of Parameters For each Models}, it is visible that Faster RCNN with VGG19 has the highest amount of trainable parameters. On the contrary, a lighter version of Faster RCNN with 1024 neurons in the FC layers of the Classifier network with MobileNet feature extractor network has the lowest trainable parameters.


\begin{table}[]
	\caption{Total number of parameters in each model}
	\label{tab:Total Number of Parameters For each Models}
\begin{tabular*}{\textwidth}{c @{\extracolsep{\fill}} cc}

\hline
Model                                                                                                           & \begin{tabular}[c]{@{}l@{}}Shared Feature Extractor Network   \\ (Pretrained with ImageNet)\end{tabular} & \begin{tabular}[c]{@{}l@{}}Total number \\ of parameters (millions) \end{tabular} \\ \hline
\multirow{3}{*}{\begin{tabular}[c]{@{}l@{}}Faster RCNN (4096 neurons in   Fast RCNN’s FC)\end{tabular}} & VGG16                                                                                                    & 139.2M                                                             \\ \cline{2-3}
                                                                                                                & VGG19                                                                                                    & 142.1M                                                            \\ \cline{2-3}
                                                                                                                & MobileNet                                                                                                & 123.7M                                                             \\ \hline
\multirow{3}{*}{\begin{tabular}[c]{@{}l@{}}Faster RCNN (1024 neurons in   Fast RCNN’s FC)\end{tabular}} & VGG16                                                                                                    & 46.2M                                                              \\ \cline{2-3}
                                                                                                                & VGG19                                                                                                    & 49.2M                                                            \\ \cline{2-3}
                                                                                                                & MobileNet                                                                                                & 30.8M                                                              \\ \hline
\multirow{3}{*}{\begin{tabular}[c]{@{}l@{}}Improved RCNN (Proposed Model)\end{tabular}}                 & VGG16                                                                                                    & 44.2M                                                           \\ \cline{2-3}
                                                                                                                & VGG19                                                                                                    & 49.5M                                                              \\ \cline{2-3}
                                                                                                                & MobileNet                                                                                                & 37.1M                                                            \\ \hline
\end{tabular*}
\end{table}

\begin{figure}[!h]
	\centering
		\includegraphics[height = 6cm, width = 16cm]{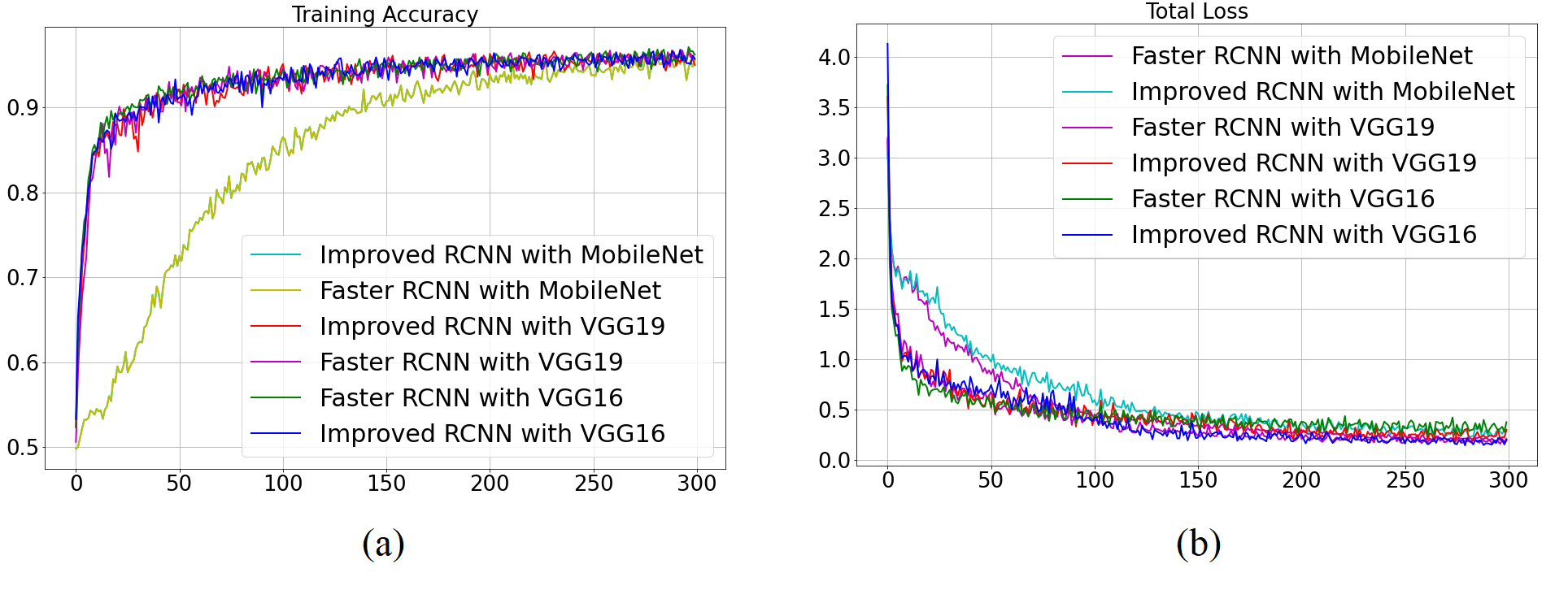}
		\caption{(a) Convergence of training accuracy, and (b) training loss per epoch}

	\label{fig:convergence}
\end{figure}

Figure~\ref{fig:convergence} depicts the convergence of training accuracy and loss over 300 epochs for Faster RCNN and our proposed Improved RCNN Model. Although having a very light structure, the convergence rate of our proposed Improved RCNN with VGG16 is almost the same as Faster RCNN with VGG16. Besides, the loss curve represents that our model produces the lowest amount of loss.

\begin{table}[]


	\centering
	\caption{Comparison among Faster RCNN with VGG16, VGG19, MobileNet as backbone and Inception based Improved RCNN with VGG16 VGG19, MobileNet as backbone}
	\label{tab:Comparison among Faster RCNN with VGG16, VGG19, MobileNet as Backbone and Inception Based Improved RCNN with VGG16 VGG19, MobileNet as Backbone}
		\small
		\small
		\resizebox{\linewidth}{!}{
		\begin{tabular}{|l|lll|lll|lll|}
			\hline
			\multirow{2}{*}{Model Vs Efficiency}       & \multicolumn{3}{c|}{\begin{tabular}[c]{@{}c@{}}Faster RCNN \\    \\ (4096 neurons in Fast RCNN’s FC)\end{tabular}} & \multicolumn{3}{c|}{\begin{tabular}[c]{@{}c@{}}Faster RCNN \\    \\ (1024 neurons in Fast RCNN’s FC)\end{tabular}} & \multicolumn{3}{c|}{Improved RCNN}                     \\ \cline{2-10}
			& \multicolumn{1}{l|}{VGG16}                  & \multicolumn{1}{l|}{VGG19}                 & MobileNet               & \multicolumn{1}{l|}{VGG16}                  & \multicolumn{1}{l|}{VGG19}                 & MobileNet               & \multicolumn{1}{l|}{VGG16}   & \multicolumn{1}{l|}{VGG19}   & MobileNet \\ \hline
			Train acc.              & \multicolumn{1}{l|}{96.17\%}                & \multicolumn{1}{l|}{95.88\%}               & 95.38\%                 & \multicolumn{1}{l|}{94.58\%}                & \multicolumn{1}{l|}{95.75\%}               & 94.88\%                 & \multicolumn{1}{l|}{95.70\%} & \multicolumn{1}{l|}{96.28\%} & 94.32\%   \\ \hline
			Train time (per epochs) & \multicolumn{1}{l|}{130.91s}                & \multicolumn{1}{l|}{138.19s}               & 108.07s                 & \multicolumn{1}{l|}{129.50s}                 & \multicolumn{1}{l|}{124.43s}               & 71.64s                  & \multicolumn{1}{l|}{120.1s}     & \multicolumn{1}{l|}{122.06s} & 94.32s    \\ \hline
			mAP(before IoU optim.)  & \multicolumn{1}{l|}{86.7\%}                 & \multicolumn{1}{l|}{91.1\%}                & 73.2                    & \multicolumn{1}{l|}{86.9\%}                 & \multicolumn{1}{l|}{86.7\%}                & 73.2\%                  & \multicolumn{1}{l|}{86.46\%} & \multicolumn{1}{l|}{91.6\%}  & 74.9\%    \\ \hline
			Test time               & \multicolumn{1}{l|}{1.81s}                  & \multicolumn{1}{l|}{1.69s}                 & 1.76s                   & \multicolumn{1}{l|}{1.71s}                  & \multicolumn{1}{l|}{1.60s}                 & 1.60s                   & \multicolumn{1}{l|}{1.68s}   & \multicolumn{1}{l|}{1.71s}   & 1.67s     \\ \hline
			True Pos.               & \multicolumn{1}{l|}{912}                    & \multicolumn{1}{l|}{894}                   & 660                     & \multicolumn{1}{l|}{902}                    & \multicolumn{1}{l|}{912}                   & 669                     & \multicolumn{1}{l|}{915}     & \multicolumn{1}{l|}{882}     & 583       \\ \hline
			False Pos.              & \multicolumn{1}{l|}{45}                     & \multicolumn{1}{l|}{57}                    & 535                     & \multicolumn{1}{l|}{44}                     & \multicolumn{1}{l|}{36}                    & 549                     & \multicolumn{1}{l|}{35}      & \multicolumn{1}{l|}{53}      & 473       \\ \hline
			False Neg.              & \multicolumn{1}{l|}{21}                     & \multicolumn{1}{l|}{49}                    & 263                     & \multicolumn{1}{l|}{31}                     & \multicolumn{1}{l|}{21}                    & 274                     & \multicolumn{1}{l|}{18}      & \multicolumn{1}{l|}{51}      & 350       \\ \hline
			Precision               & \multicolumn{1}{l|}{95\%}                   & \multicolumn{1}{l|}{94\%}                  & 56\%                    & \multicolumn{1}{l|}{95\%}                   & \multicolumn{1}{l|}{96.20\%}               & 55\%                    & \multicolumn{1}{l|}{96.31\%} & \multicolumn{1}{l|}{92\%}    & 55\%      \\ \hline
			Recall                  & \multicolumn{1}{l|}{98\%}                   & \multicolumn{1}{l|}{96\%}                  & 72\%                    & \multicolumn{1}{l|}{97\%}                   & \multicolumn{1}{l|}{97.75\%}               & 71\%                    & \multicolumn{1}{l|}{98.07\%} & \multicolumn{1}{l|}{96\%}    & 62\%      \\ \hline
			F2 Score                & \multicolumn{1}{l|}{97\%}                   & \multicolumn{1}{l|}{95\%}                  & 68\%                    & \multicolumn{1}{l|}{96\%}                   & \multicolumn{1}{l|}{97.44\%}               & 66\%                    & \multicolumn{1}{l|}{97.71\%} & \multicolumn{1}{l|}{94\%}    & 60\%      \\ \hline
		\end{tabular}}

\end{table}
We compare the performance of these models and show the tradeoff between the model size and performance. For this purpose, we use 934 images. From Table~\ref{tab:Comparison among Faster RCNN with VGG16, VGG19, MobileNet as Backbone and Inception Based Improved RCNN with VGG16 VGG19, MobileNet as Backbone} we can see that the training time for our prposed model is the lowest but the testing time is slightly higher than Faster RCNN with FC layers having 1024 neurons. Because in the latter case the number of fully connected layers is one-fourth and it reduces the size of the parameters. The training and testing accuracy is highest for our Improved RCNN with VGG19 backbone.

\begin{table}[]
	\centering
	\caption{Classwise Performance for each classes}
	\label{tab:Classwise Performance for each classes}
		\small
		\small
\resizebox{\linewidth}{!}{
		\begin{tabular}{|ll|lll|lll|llll|}
			\hline
			\multicolumn{2}{|l|}{\multirow{2}{*}{\begin{tabular}[c]{@{}l@{}}Model Vs Class\end{tabular}}} & \multicolumn{3}{l|}{\begin{tabular}[c]{@{}l@{}}Faster RCNN \\    \\ (4096 neurons in   Fast RCNN’s FC)\end{tabular}} & \multicolumn{3}{l|}{\begin{tabular}[c]{@{}l@{}}Faster RCNN \\    \\ (1024 neurons in   Fast RCNN’s FC)\end{tabular}} & \multicolumn{4}{l|}{Improved RCNN (Proposed Method)}                                      \\ \cline{3-12}
			\multicolumn{2}{|l|}{}                                                                                                 & \multicolumn{1}{l|}{VGG16}                  & \multicolumn{1}{l|}{VGG19}                 & MobileNet                 & \multicolumn{1}{l|}{VGG16}                  & \multicolumn{1}{l|}{VGG19}                 & MobileNet                 & \multicolumn{1}{l|}{VGG16} & \multicolumn{1}{l|}{VGG19} & \multicolumn{2}{l|}{MobileNet} \\ \hline
			\multicolumn{2}{|l|}{Early blight}                                                                                     & \multicolumn{1}{l|}{0.95}                   & \multicolumn{1}{l|}{0.82}                  & 0.63                      & \multicolumn{1}{l|}{0.87}                   & \multicolumn{1}{l|}{0.93}                  & 0.55                      & \multicolumn{1}{l|}{0.93}  & \multicolumn{1}{l|}{0.73}  & \multicolumn{2}{l|}{0.42}      \\ \hline
			\multicolumn{2}{|l|}{Two spotted spider   mite}                                                                        & \multicolumn{1}{l|}{0.89}                   & \multicolumn{1}{l|}{0.98}                  & 0.49                      & \multicolumn{1}{l|}{0.97}                   & \multicolumn{1}{l|}{0.89}                  & 0.59                      & \multicolumn{1}{l|}{0.94}  & \multicolumn{1}{l|}{1.0}   & \multicolumn{2}{l|}{0.63}      \\ \hline
			\multicolumn{2}{|l|}{Healthy}                                                                                          & \multicolumn{1}{l|}{0.95}                   & \multicolumn{1}{l|}{0.98}                  & 0.84                      & \multicolumn{1}{l|}{0.99}                   & \multicolumn{1}{l|}{0.99}                  & 0.82                      & \multicolumn{1}{l|}{0.99}  & \multicolumn{1}{l|}{0.99}  & \multicolumn{2}{l|}{0.96}      \\ \hline
			\multicolumn{2}{|l|}{Late  blight}                                                                                      & \multicolumn{1}{l|}{0.90}                   & \multicolumn{1}{l|}{0.95}                  & 0.43                      & \multicolumn{1}{l|}{0.93}                   & \multicolumn{1}{l|}{0.95}                  & 0.38                      & \multicolumn{1}{l|}{0.99}  & \multicolumn{1}{l|}{0.97}  & \multicolumn{2}{l|}{0.41}      \\ \hline
			\multicolumn{2}{|l|}{Target Spot}                                                                                      & \multicolumn{1}{l|}{0.95}                   & \multicolumn{1}{l|}{0.92}                  & 0.5                       & \multicolumn{1}{l|}{0.99}                   & \multicolumn{1}{l|}{0.99}                  & 0.43                      & \multicolumn{1}{l|}{0.96}  & \multicolumn{1}{l|}{0.91}  & \multicolumn{2}{l|}{0.43}      \\ \hline
			\multicolumn{2}{|l|}{Bacterial spot}                                                                                   & \multicolumn{1}{l|}{0.98}                   & \multicolumn{1}{l|}{0.95}                  & 0.45                      & \multicolumn{1}{l|}{0.96}                   & \multicolumn{1}{l|}{0.96}                  & 0.44                      & \multicolumn{1}{l|}{0.96}  & \multicolumn{1}{l|}{0.94}  & \multicolumn{2}{l|}{0.0}       \\ \hline
			\multicolumn{2}{|l|}{Leaf mold}                                                                                        & \multicolumn{1}{l|}{0.99}                   & \multicolumn{1}{l|}{0.96}                  & 0.58                      & \multicolumn{1}{l|}{0.95}                   & \multicolumn{1}{l|}{0.99}                  & 0.63                      & \multicolumn{1}{l|}{0.99}  & \multicolumn{1}{l|}{0.98}  & \multicolumn{2}{l|}{0.72}      \\ \hline
			\multicolumn{2}{|l|}{Yellow Leaf Curl   Virus}                                                                         & \multicolumn{1}{l|}{0.99}                   & \multicolumn{1}{l|}{0.94}                  & 0.75                      & \multicolumn{1}{l|}{0.97}                   & \multicolumn{1}{l|}{0.99}                  & 0.9                       & \multicolumn{1}{l|}{0.99}  & \multicolumn{1}{l|}{0.95}  & \multicolumn{2}{l|}{0.79}      \\ \hline
			\multicolumn{2}{|l|}{Septoria leaf spot}                                                                               & \multicolumn{1}{l|}{0.96}                   & \multicolumn{1}{l|}{0.97}                  & 0.5                       & \multicolumn{1}{l|}{0.95}                   & \multicolumn{1}{l|}{0.98}                  & 0.41                      & \multicolumn{1}{l|}{0.94}  & \multicolumn{1}{l|}{0.92}  & \multicolumn{2}{l|}{0.45}      \\ \hline
			\multicolumn{2}{|l|}{Mosaic virus}                                                                                     & \multicolumn{1}{l|}{1.00}                   & \multicolumn{1}{l|}{1.00}                  & 0.71                      & \multicolumn{1}{l|}{1.00}                   & \multicolumn{1}{l|}{1.00}                  & 0.71                      & \multicolumn{1}{l|}{0.95}  & \multicolumn{1}{l|}{0.95}  & \multicolumn{2}{l|}{0.76}      \\ \hline
			\multicolumn{2}{|l|}{mAP}                                                                                              & \multicolumn{1}{l|}{0.95}                   & \multicolumn{1}{l|}{0.94}                  & 0.56                      & \multicolumn{1}{l|}{0.95}                   & \multicolumn{1}{l|}{0.96}                  & 0.55                      & \multicolumn{1}{l|}{0.963}  & \multicolumn{1}{l|}{0.92}  & \multicolumn{2}{l|}{0.55}      \\ \hline
		\end{tabular}}
\end{table}

Interestingly, after applying IoU optimization the testing efficiency of Improved RCNN with VGG16 backbone network outperforms all the others by having the highest True Positive ($TP$) and lowest False Positive ($FP$) and False Negative ($FN$). The higher number of $TP$ and lower number of $FP$ and $FN$ instances yields a smaller amount of misclassification for our model in terms of precision, recall, and F2 Score. From Table~\ref{tab:Comparison among Faster RCNN with VGG16, VGG19, MobileNet as Backbone and Inception Based Improved RCNN with VGG16 VGG19, MobileNet as Backbone} it is visible that our Improved RCNN model with VGG16 backbone network performs much better than the others in terms of Precision, F2 Score, and Recall. Class-wise precision for each of the considered models is listed in Table~\ref{tab:Classwise Performance for each classes}. For each class, the best accuracy is highlighted. Our proposed Improved RCNN with the VGG16 base network achieves the best result for most of the cases.

For Two-Spotted Spider Mites, Healthy and Late Blight diseases, the proposed Improved RCNN with VGG19 base works slightly better. For Early Blight, Bacterial Spot, and Mosaic Virus diseases, Faster RCNN with VGG16 base works slightly better. However for both of these models the number of trainable parameters is higher than the Improved RCNN with VGG19 base.

\begin{figure}
     \centering
     \begin{subfigure}[b]{0.3\textwidth}
         \centering
         \includegraphics[width=\textwidth]{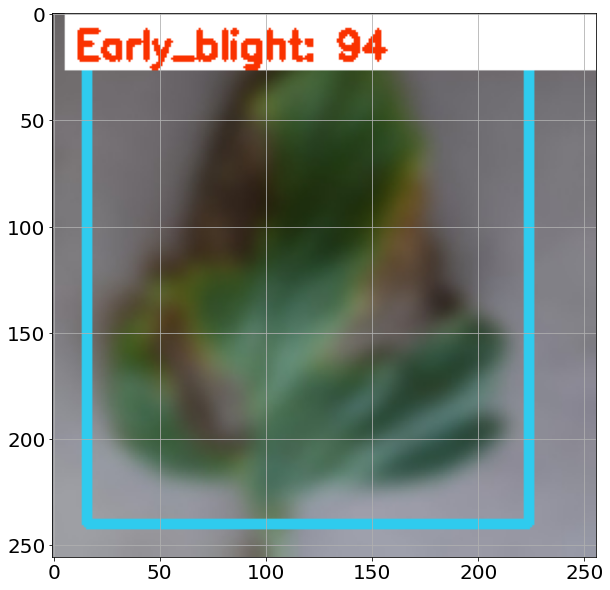}
         \caption{Correctly classified}
         \label{fig:(a)}
     \end{subfigure}
     \hfill
     \begin{subfigure}[b]{0.3\textwidth}
         \centering
         \includegraphics[width=\textwidth]{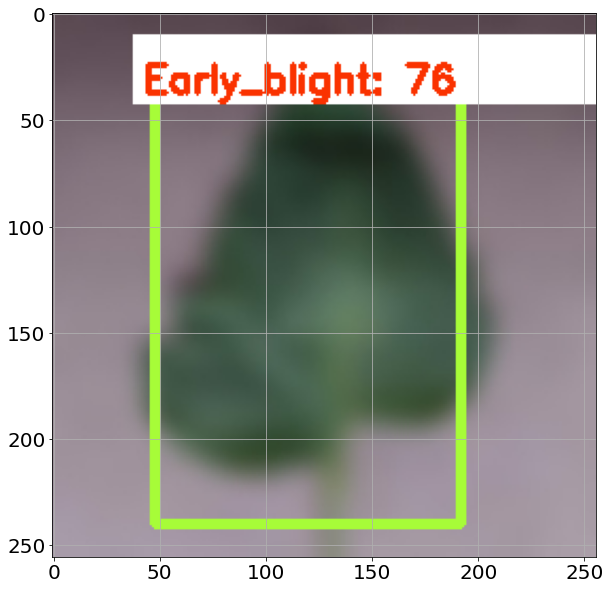}
         \caption{Miss-classified}
         \label{fig:(b)}
     \end{subfigure}
     \hfill
     \begin{subfigure}[b]{0.3\textwidth}
         \centering
         \includegraphics[width=\textwidth]{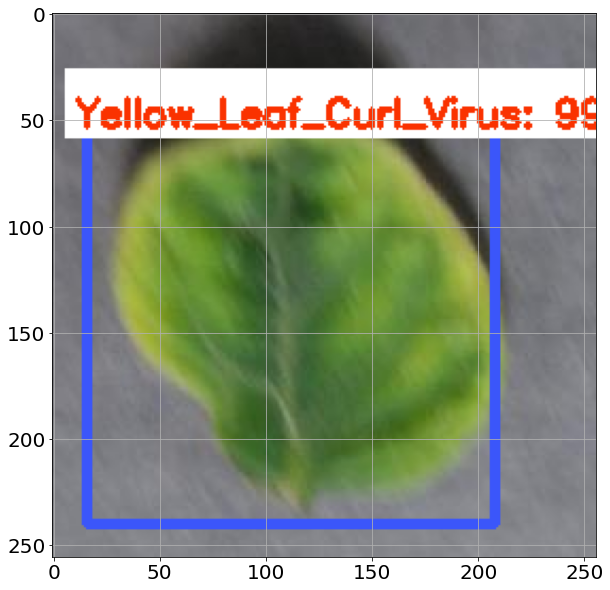}
         \caption{Correctly classified}
         \label{fig:(c)}
     \end{subfigure}
     \begin{subfigure}[b]{0.3\textwidth}
         \centering
         \includegraphics[width=\textwidth]{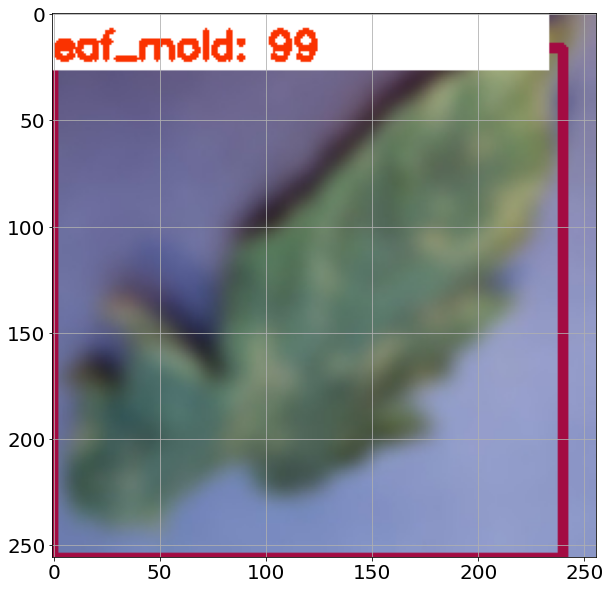}
         \caption{Correctly classified}
         \label{fig:(d)}
     \end{subfigure}
     \hfill
     \begin{subfigure}[b]{0.3\textwidth}
         \centering
         \includegraphics[width=\textwidth]{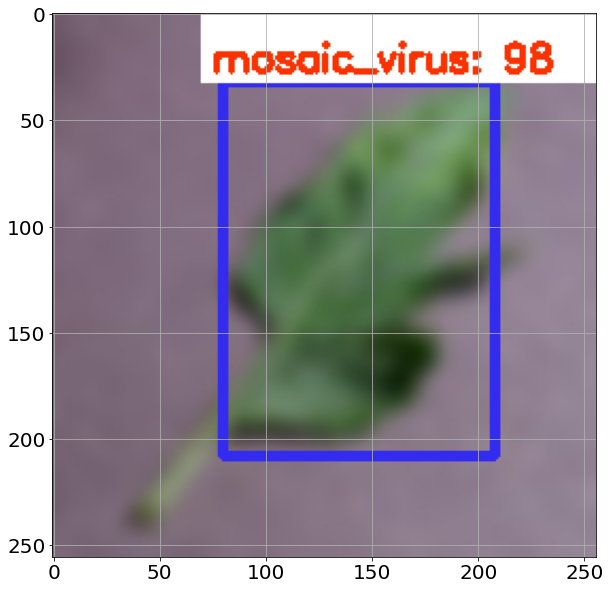}
         \caption{Correctly classified}
         \label{fig:(e)}
     \end{subfigure}
     \hfill
     \begin{subfigure}[b]{0.3\textwidth}
         \centering
         \includegraphics[width=\textwidth]{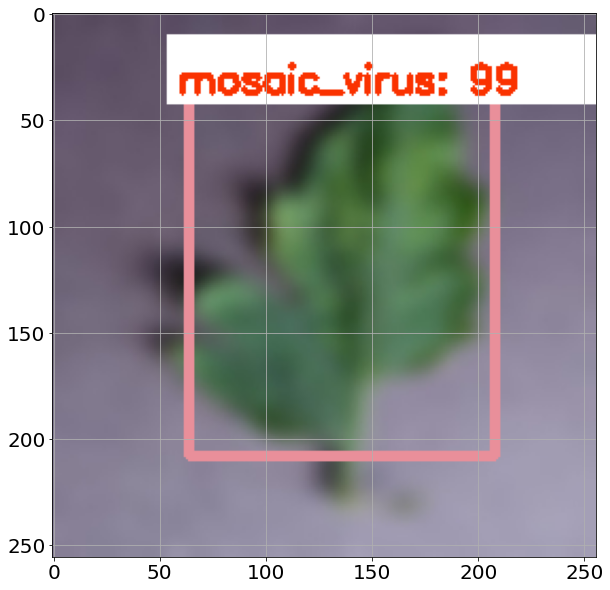}
         \caption{Correctly classified}
         \label{fig:(f)}
     \end{subfigure}
        \caption{Testing result of our proposed Improved RCNN model for blur images}
        \label{fig:blur test}
\end{figure}

Our motivation is to implement a deep learning model that will be suitable for implementation in real farming. That is why we take into consideration the fact that the images captured by farmers may be blurred sometimes due to shaky hand, low configuration smart phone camera, and some other environmental factors. To test the reliability of our model in such situation we extract 100 blurred images from the dataset and evaluate our proposed algorithm using it. Figure~\ref{fig:blur test} illustrates the detection result of our model.  In this figure, the bounding boxes detecting the leaves are quite accurate. However, the image of Figure~\ref{fig:(b)} is a miss-classified result due to being extremely blurred. The other five images are moderately blurred and hence they are correctly classified. Table~\ref{tab:Classwise Evaluation Result of the Proposed IRCNN model for Blurred Images} states the class-wise evaluation result of our proposed model from the blur dataset. We see that our model cannot successfully classify the images from Target Spot, Septoria Leaf Spot. To compare performance of other algorithms with that of ours for blur images, we apply Faster RCNN with VGG16, VGG19 and MobileNet Feature extractor for the blurry dataset.  Table~\ref{tab:Comparison Between Faster RCNN and Improved RCNN for Blurred Images} shows the class-wise evaluation result of our proposed model and other models for the blur images. We can see that for other models precision, recall and F2 score are below 50\% which is remarkably low. So the results demonstrate that our proposed model achieves better accuracy than other models on blur images.

\begin{table}[]
\caption{Class-wise evaluation result of the proposed Improved RCNN model for blurred images}
\label{tab:Classwise Evaluation Result of the Proposed IRCNN model for Blurred Images}
\begin{tabular*}{\textwidth}{c @{\extracolsep{\fill}} cccccc}
\hline
Class Name             & TP & FP & FN & F2 Sores & Precision & Recall \\ \hline
Early Blight           & 9  & 20 & 1  & 0.65     & 0.31      & 0.9    \\
Spider Mites           & 2  & 0  & 8  & 0.24     & 1.00      & 0.2    \\
Healthy                & 10 & 0  & 0  & 1.00     & 1.00      & 1.0    \\
Late Blight            & 8  & 18 & 2  & 0.60     & 0.31      & 0.8    \\
Target Spot            & 0  & 0  & 10 & 0.00     & 0.00      & 0.0    \\
Bacterial Spot         & 6  & 0  & 4  & 0.65     & 1.00      & 0.6    \\
Leaf Mold              & 7  & 0  & 3  & 0.74     & 1.00      & 0.7    \\
Yellow Leaf Curl Virus & 10 & 5  & 0  & 0.91     & 0.67      & 1.0    \\
Septoria Leaf Spot     & 0  & 0  & 10 & 0.00     & 0.00      & 0.0    \\
Mosaic Virus           & 8  & 1  & 2  & 0.81     & 0.89      & 0.8    \\ \hline
Overall                & 60 & 44 & 40 & 0.59     & 0.58      & 0.6    \\ \hline
\end{tabular*}
\end{table}

\begin{table}[]
\caption{Comparison between Faster RCNN and Improved RCNN for Blurred Images}
\label{tab:Comparison Between Faster RCNN and Improved RCNN for Blurred Images}
\begin{tabular*}{\textwidth}{c @{\extracolsep{\fill}} cccccc}
\hline
Model                  & Precision & Recall & F2 Score \\ \hline
Improved RCNN with VGG16            & 0.58      & 0.60    & 0.59\\
Faster RCNN VGG16            & 0.33      & 0.41   & 0.39\\
Improved RCNN with VGG19            & 0.49      & 0.46   & 0.46\\
Faster RCNN with VGG19            & 0.42      & 0.54   & 0.51\\
Improved RCNN with MobileNet        & 0.41      & 0.11   & 0.13\\
Faster RCNN with MobileNet        & 0.53      & 0.16   & 0.19\\ \hline

\end{tabular*}
\end{table}

\begin{figure}[!h]
	\centering
	\includegraphics[height = 8cm, width = 10cm]{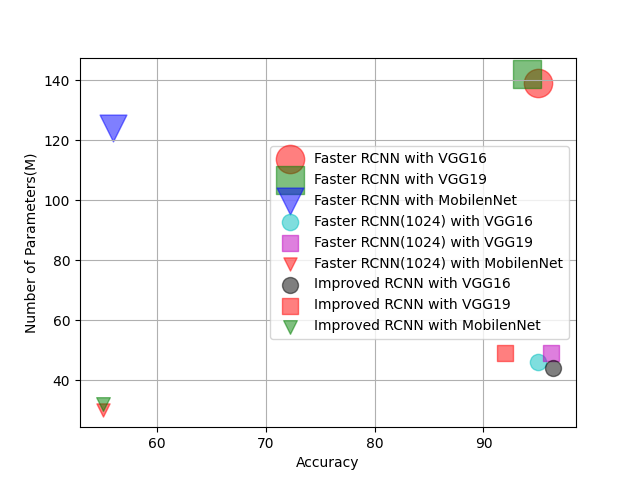}
	\caption{Number of Parameters VS accuracy}
	\label{fig:Number of Parameters VS accuracy}
\end{figure}

Figure~\ref{fig:Number of Parameters VS accuracy} illustrates a visual comparison between the numbers of trainable parameters versus achieved accuracy for each of these models. From the figure, it is visible that a lighter version of Faster RCNN (1024 Neurons in FC layers) with MobileNet and Improved RCNN with MobileNet has the lowest number of parameters, but their performance is lower in terms of accuracy. On the other hand, Improved RCNN with VGG16 has the third-lowest number of parameters, still yields the best performance in terms of accuracy.\\

\textbf{Summary of Findings: }
From our detailed analysis of the experimental results discussed above we have some substantial findings which are listed below:
\begin{itemize}
\item After applying the bounding box optimization the accuracy of our model is found to be quite good. Confusion matrix depicts a very tiny amount of misclassification. For most of the disease classes the detection rate of our model is outstanding.
\item Our proposed model converges faster than existing baseline models within the first few epochs with a satisfactory level of training accuracy.
\item The training and testing time required for the proposed model is acceptable as compared to the baselines.
\item Our model has very small number of trainable parameters compared to the baseline model (namely, Faster RCNN). It thus offers a relatively lightweight architecture that is suitable for real life usage. Despite having small number of parameters, our model surpasses the other models in terms of mean average precision, recall, and F2-score.
\item Our model is able to detect diseased leafs from blur images in a better way than the baseline models. This is very promising when it comes to implementing models in field-level agriculture.
\end{itemize}


\section{Conclusion}
\label{sec:conclusion}
The involvement of deep learning in agricultural image processing domain has a promising future for automated farming. Numerous researches have taken place for classifying plant diseases from leaf images. But localization of disease-affected leaves from images is still a growing research area that needs more attention. 
Our research focuses on design of an improved RCNN algorithm for the detection, localization, and classification of tomato leaf disease. Our proposed model  improves the RCNN algorithm by changing the fully connected classifier model to an inception-based classifier model. We have used VGG16 as the base network for performing deep feature extraction. For better detection performance, we have applied IoU optimization. The experimental evaluation shows that our model can correctly localize and recognize tomato leaf diseases efficiently with an mAP of 96.31\%. The recall and F2 scores are 97.71\% and 98.07\% respectively. Our proposed model can easily be deployed in a larger system where drones take images of leaves and these images will be fed into our model to know the health condition.

This research spawns a number of interesting directions. We may use some other pre-trained feature extractor networks and perform comparison among them in terms of accuracy, precision, memory usage, and time complexity. Collecting some field-level data and applying the model on that dataset may increase the credibility of the model. Implementing an embedded system based on our model may assist the need for on-site inspection by experts.

\section*{Acknowledgement}
Hasin Rehana is supported by a Research Fellowship funded by the Information and Communication Technology Division, Ministry of Telecommunications and Information Technology, Government of Bangladesh.



\bibliography{cas-refs}
\bibliographystyle{unsrt}





\end{document}